\documentclass{llncs}
\usepackage{hyperref}

\makeatletter
\newcommand{\printfnsymbol}[1]{%
  \textsuperscript{\@fnsymbol{#1}}%
}
\makeatother

\usepackage{makeidx}  
\usepackage[utf8]{inputenc}

\usepackage{graphicx}
\usepackage{caption}
\usepackage{subcaption}
\usepackage{nomencl}
\makenomenclature

\usepackage{tabularx,algorithm,algorithmic,units,rotating}

\begin{document}

\title{Detecting Potential Local Adversarial Examples for Human-Interpretable Defense}
\titlerunning{Detecting Potential Local Adversarial Examples}  
%
\author{Xavier Renard\inst{1}\thanks{Equal contribution} \and Thibault Laugel\inst{2}\printfnsymbol{1}
Marie-Jeanne Lesot\inst{2} \and Christophe Marsala\inst{2} \and Marcin Detyniecki\inst{1,2,3}}
\authorrunning{Xavier Renard et al.} 
%
\tocauthor{Xavier Renard, Thibault Laugel, Marie-Jeanne Lesot, Christophe Marsala and Marcin Detyniecki}
\institute{AXA Group, Paris, France\\
\email{xavier.renard@axa.com},
\and
Sorbonne  Universit\'{e}, CNRS, Laboratoire d'Informatique de Paris 6, LIP6, F-75005 Paris, France\\
\email{thibault.laugel@lip6.fr},
\and
Polish Academy of Science, IBS PAN, Warsaw, Poland}

\maketitle

\begin{abstract}
Machine learning models are increasingly used in the industry to make decisions such as credit insurance approval. Some people may be tempted to manipulate specific variables, such as the age or the salary, in order to get better chances of approval. In this ongoing work, we propose to discuss, with a first proposition, the issue of detecting a potential local adversarial example on classical tabular data by providing to a human expert the locally critical features for the classifier's decision, in order to control the provided information and avoid a fraud.
\end{abstract}
\section{Introduction}

As machine learning models are increasingly being used to make decisions that directly impact people, the risk of fraudulent attempts to fool these models is more present than ever.
Let us consider the case of an insurance company with an automated service that allows customers to apply for credit insurance. Through a dedicated interface, clients have to provide personal details so a machine learning prediction model can assess their risks of default and decide automatically whether they are eligible for the insurance.
A potential customer may want to improve his chances of getting the credit insurance and could be tempted to game the system by fooling the decision model.
The malicious customer could lie to some of the questions he is being asked, such as his salary or age.
These alterations will affect the input provided to the automatic classifier and may change the prediction.
Such attacks could provoke major issues for an insurance company: an inaccurate risk assessment can lead to a possibility of bankruptcy.
Then, how is it possible to prevent these attacks, and more precisely, what defense strategy should be applied against these attacks?

In this paper, we address the following issue: 
given an automated classifier and a potential customer, our objective is to identify a sparse set of features that are the most important for the classifier's prediction made for this customer and thus the ones that could easily alter the prediction.
Given these most important features, it is possible to ask the potential client for additional information or proof, in order to counter a potential adversarial attack by ascertaining the truthfulness of his declarations.
The proposed approach is framed at the intersection of adversarial machine learning and human-interpretable machine learning. Given a point in a feature space (\emph{ie.} a potential client) we propose to search the closest decision boundaries of the classifier, which we assume to be the most relevant for the point's prediction, and fit an interpretable surrogate model to approximate these boundaries. The most important features for this prediction are then extracted from the surrogate.
These features lead to checks in order to prevent a potential adversarial attack.
This defense approach is local (\emph{ie} centered around the prediction to assess), focused on classical feature-based classification problems and model-agnostic (\emph{i.e.} independent to the choice of the black-box classifier), since we assume that no information about the classifier is made available for the proposed approach.

The next section details the relevant background for this work. Section~\ref{sec:proposition} presents a formalization of the considered problem and a description of the principles of our proposition. Section~\ref{sec:experiment} proposes a first evaluation of our proposition on the German Credit dataset: we quantify the fidelity to the classifier and we provide an example of explanation that can be used as a defense against an attack by clarifying whether an input is an adversarial example or not.
We conclude this paper by outlining the perspectives to pursue this work.

\section{Background}

In adversarial machine learning, evasion attacks~\cite{Biggio2013,Goodfellow2015,Papernot2016limitations} aim at generating adversarial examples that fool a machine learning classifier. Defense techniques have been developed to counter these attacks: either by changing the training process of the classifier to be attacked to make it more robust to small perturbations~\cite{Papernot2017practical,Bhagoji2017} or by adding a detection mechanism to identify potential adversarial examples~\cite{Cao2017}. However, both strategies generally rely on strong assumptions or knowledge, either about the dataset or about our capacity to identify the \textit{true} label of an instance. For instance, when dealing with images, a human is able to provide at test time the true label despite an adversarial perturbation. In the mentioned insurance context, the classifier's output for a potential adversarial example can't be compared with the true label since it is unknown (eg. a default would happen in the future). To mitigate this issue, we look into the classifier's inner working to assess the potential weaknesses of a client's application, and control the veracity of problematic information he may have provided.

To do so, the field of machine learning interpretability aims at generating explanations to provide insights about a prediction made by a black-box classifier. Explanations can take the form of feature importance vectors that quantify the relative impact of each feature in the prediction~\cite{Baehrens2010,Ribeiro2016,Lundberg2016}. To extract these feature importances, we focus on surrogate model approaches which attempt to approximate the decision boundary of a black-box classifier with a simpler \textit{interpretable} model either at global scale (to mimic the global behavior of a black-box model)~\cite{Craven1996,Hara2016a} or at local scale (to mimic the black-box behavior locally, around a prediction for instance)~\cite{Ribeiro2016,laugel2018}. 
Surrogate model approaches can somewhat be related to the task of model theft in adversarial learning, which consists in building a substitute model to copycat the decisions of a black-box classifier, either to steal it and use its outputs~\cite{Tramr2016StealingML} or to use it for transferable evasion attacks~\cite{Papernot2017practical}.

\section{Proposition: Local Adversarial Detection (LAD)}\label{sec:proposition}



%
%

The following context is considered: a black-box classifier $b:\mathcal{X}\rightarrow \mathcal{Y}$ is trained on a dataset composed of an input feature space $\mathcal{X} = \mathcal{R}^D$ made of classical features and an output $Y$ where each instance $x \in \mathcal{X}$ is associated with a class label $y \in \mathcal{Y}$.
The classifier $b$ is publicly available for querying and accepts an input vector $x$ (e.g. containing information such as age and income) to provide predictions $b(x)$ that lead to business decisions (such as insurance or credit acceptance).
For security reasons, each input $x$ leading to a prediction $b(x)$ is considered suspicious: it is considered as being a potential adversarial example $\hat{x}$ (\emph{i.e.} fake information), generated to get a more favourable decision from the black-box classifier $b$ such that $b(x) \neq b(\hat{x})$.
In the context of insurance, it is assumed that the provided input data is reasonable since it deals with real or physical variables.

In this work, our objective is to design a 
security layer to provide an adaptive defense that returns the most important features to check in order to clarify whether or not an input $\hat{x}$ should be investigated.
Given a prediction $b(\hat{x})$ and a potential adversarial example $\hat{x}$ to assess, we aim at providing the locally most important features for the black-box classifier that lead to the prediction $b(\hat{x})$.
These features are also the most sensitive for someone willing to fool the classifier~$b$, and can be investigated by a human expert or a machine to ask for a relevant set of proofs to avoid a fraud.
For instance, given $\hat{x}$, if the income is the only important feature that would significantly impacts the prediction $b(\hat{x})$, then a single proof of income would be necessary to counter an adversarial attack. Our objective is thus to identify a set $\mathcal{F}_{\hat{x}}$ of features from $\mathcal{X}$ that are key for the prediction $b(\hat{x})$ and should be checked to clarify whether $\hat{x}$ is a malicious customer or not.

\begin{figure}[t]
	\centering
	
	\begin{subfigure}{0.31\textwidth}
		\includegraphics[width=\textwidth]{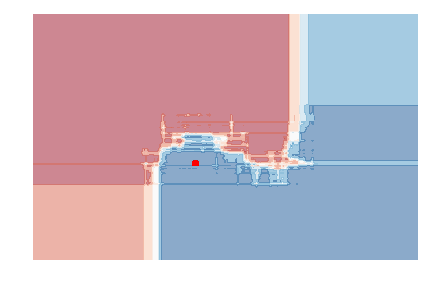}
		\caption{Potential adversarial example $\hat{x}$ (red dot) over black-box classifier's boundaries}
		\label{fig:algoa}
	\end{subfigure}
	~
	\begin{subfigure}{0.31\textwidth}
		\includegraphics[width=\textwidth]{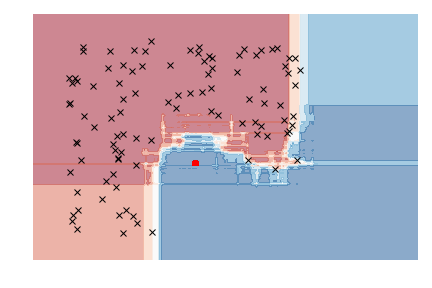}
		\caption{Random generation of support points $x_{sp}$ where $b(x_{sp}) \neq b(\hat{x})$}
		\label{fig:algob}
	\end{subfigure}
	~
	\begin{subfigure}{0.31\textwidth}
		\includegraphics[width=\textwidth]{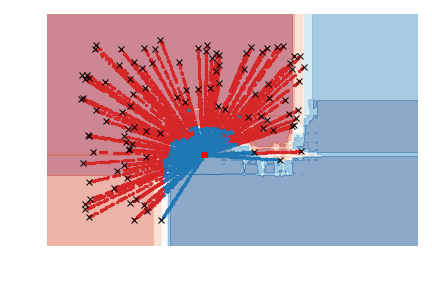}
		\caption{Random generation and classification of points on segments $[\hat{x};x_{sp}]$}
		\label{fig:algoc}
	\end{subfigure}
	
	\begin{subfigure}{0.31\textwidth}
		\includegraphics[width=\textwidth]{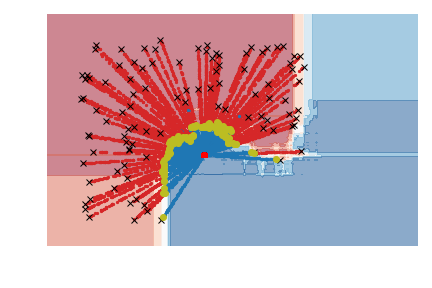}
		\caption{Boundary search: segment points that maximize the information gain}
		\label{fig:algod}
	\end{subfigure}
	~
	\begin{subfigure}{0.31\textwidth}
		\includegraphics[width=\textwidth]{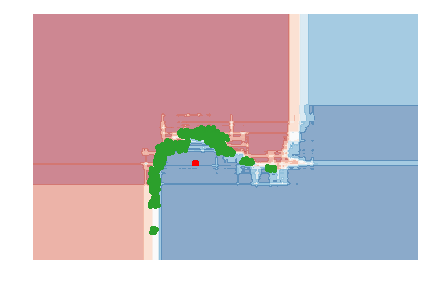}
		\caption{Generation and classification of points around the boundary touchpoints}
		\label{fig:algoe}
	\end{subfigure}
	~
	\begin{subfigure}{0.31\textwidth}
		\includegraphics[width=\textwidth]{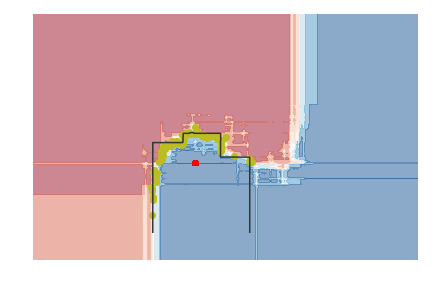}
		\caption{Boundaries of the trained interpretable surrogate}
		\label{fig:algof}
	\end{subfigure}
	
	\caption{Principle of Local Adverse Detection (LAD)}
	\label{fig:algo}
\end{figure}

To solve that problem, we propose to fit a local surrogate in order to approximate the closest local black-box boundaries to the potential adversarial example $\hat{x}$. Then, the set $\mathcal{F}_{\hat{x}}$ of features that are key for the prediction $b(\hat{x})$ is extracted from the local surrogate. 
The proposed Algorithm~\ref{algo} and its different steps illustrated Figure~\ref{fig:algo} work as follows.
To detect the closest decision boundaries, $N$ support points $x_{sp}^i$ are drawn in $\mathcal{X}$ with $b(x_{sp}) \neq b(\hat{x})$ (Figure~\ref{fig:algob}). These support points delimit segments $[\hat{x} ; x_{sp}^i]$~$ \forall i \in [1;N]$ on which the local black-box boundary should be sought: to do so, the maximum of the information gain is sought on every segment $[\hat{x} ; x_{sp}^i]$ based on $M$ points drawn on these segments then labeled using $b$ (Figures~\ref{fig:algoc} and~\ref{fig:algod}). The point on each segment where the information gain is maximal is called a boundary touchpoint $x_{bt}$. These points are on the black-box boundary or close to it: it is possible to outline locally the black-box boundary shape.
Then, an interpretable local surrogate classifier $s_{\hat{x}}$ is trained on a small set of points generated in the immediate neighbourhood of the local black-box boundary (Figures~\ref{fig:algoe} and~\ref{fig:algof}), outlined by the boundary touchpoints $x_{bt}$. Finally, the set of key features $\mathcal{F}_{\hat{x}}$ for the prediction $b(\hat{x})$ is extracted from the surrogate $s_{\hat{x}}$ that approximates the local boundary of the classifier $b$.


\begin{algorithm}[t]
\begin{algorithmic}
   \STATE {\bfseries Input:} potential adversarial example $\hat{x} \in \mathcal{X}$, classifier $b:\mathcal{X}\rightarrow\mathcal{Y}$, untrained surrogate $s_x$, $N$, $M$
   
   \STATE $x_{sp} \leftarrow$ Draw $N$ support points $x_{sp}^i \in \mathcal{X}, i \in [1...N]$ with $b(x_{sp}^i) \neq b(\hat{x})$
   \FORALL{$x_{sp}^i \in x_{sp}$}
    \STATE $x_{seg}^i \leftarrow$ Draw $M$ segment points $x_{seg}^{i,j}, j \in [1...M]$ with $x_{seg}^{i,j} \in [\hat{x};x_{seg}^i]$
    \STATE $y_{seg}^{i} \leftarrow b(x_{seg}^{i})$ label every segment points with the classifier $b$
    \STATE $x_{bt}^{i} \leftarrow$ Find the touchpoint $x_{bt}^{i} \in [\hat{x};x_{seg}^i]$ that maximizes the info gain $IG(y_{seg}^{i})$
    \STATE $X_{s_x^i} \leftarrow$ Draw $P$ points in an hypersphere of radius $r_{s_x}$ around the touchpoint $x_{bt}^{i}$
    \STATE $Y_{s_x^i} \leftarrow$ Label every points $X_{s_x^i}$ with the classifier $b$
   \ENDFOR
   \STATE $Y_{s_x} \leftarrow b(X_{s_x})$
   \STATE Train the local surrogate $s_x$ on ($X_{s_x}, Y_{s_x}$)
   \STATE $\mathcal{F}_{\hat{x}} \leftarrow$ Extract the most important features from $s_x$
   \STATE {\bfseries Return:} $\mathcal{F}_{\hat{x}}$
\end{algorithmic}
\caption{Outline of Local Adverse Detection (LAD) algorithm}
\label{algo}
\end{algorithm}
\vspace{-1.5em}

\section{Case Study: German Credit Dataset}~\label{sec:experiment}

We present a first use case towards designing a complete and robust experimental protocol to evaluate the quality of the proposed approach. This work is still in progress and requires further discussion.

We apply the proposed approach to the German Credit dataset available from UCI. A classifier $b$ (a Random Forest with 200 estimators) is trained on 70\% of the data and acts as the automated decision model accepting or rejecting the customer's application.
Considering information provided by customers $\hat{x}$ from the remaining test dataset (250 instances), the LAD algorithm is used (with $N=1000$, $M=100$ and the surrogate being a decision tree with a maximum tree depth of 5) to extract the key features for prediction $b(\hat{x})$. They constitute the features an expert should check to ascertain the prediction.

Our first experiment assesses the accurate approximation of the black-box classifier's local decision boundary by the local surrogate trained with LAD. The \emph{Local Fidelity} metric described in~\cite{laugel2018} is used to assess locally, around the potential adversarial example, the fidelity of the surrogate $s_{\hat{x}}$ to the classifier $b$.
The \emph{Local Fidelity} metric is defined as the fidelity of the local surrogate $s_{\hat{x}}$ to the black-box classifier $b$ within a neighborhood $\mathcal{V}_{\hat{x}}$ around $\hat{x}$.
$|\mathcal{V}_x|$ points are therefore drawn uniformly in the neighbourhood $\mathcal{V}_{\hat{x}}$ of $\hat{x}$ bounded by an hypersphere of radius $r$. Then, to get the local fidelity of $s_{\hat{x}}$ to $b$, the classification accuracy is computed on these points labeled with both $s_{\hat{x}}$ and $b$, such as:

\begin{equation}
    LocalFid(\hat{x}, s_{\hat{x}}) = Acc_{x_i \in \mathcal{V}_{\hat{x}}}(b(x_i), s_{\hat{x}}(x_i))
\end{equation}

\begin{figure}[t]
	\center
	\includegraphics[width=0.75\textwidth]{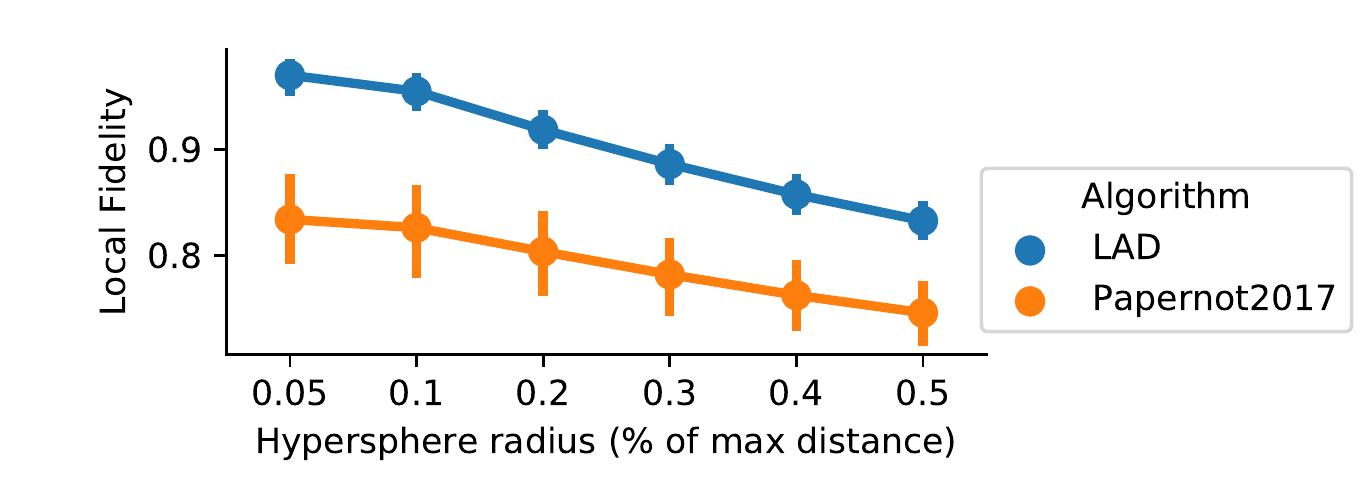}
	\caption{Local Fidelity for LAD and Papernot2017~\cite{Papernot2017practical} on German Credit dataset for several values of $r$}
	\label{fig:localfidelity_german}
\end{figure}

We set $|\mathcal{V}_x|=1000$ and the radius of the hyperspheres as a percentage (between $0.05\%$ and $0.5\%$) of the maximum distance between $\hat{x}$ and the test dataset.
To challenge the proposed approach, we choose to use the substitute model method from \cite{Papernot2017practical}, based on data generated using Jacobian augmentation.
The average local fidelity values over all the test dataset are shown Figure~\ref{fig:localfidelity_german}.
As mentioned earlier, an important difference of \cite{Papernot2017practical} compared to the proposed approach lies in the fact that the trained substitute is global instead of focusing on a specific region. Then, as shown Figure~\ref{fig:localfidelity_german}, the average local fidelity of LAD is higher for smaller radius values as expected. However, as the radius increases and the evaluation of the fidelity of the substitutes is performed on a wider scale, the difference vanishes.

An important contribution of this proposition consists in the generation of explanations given a prediction for a potential adversarial example. A very first illustration is provided in Table~\ref{table:adverse_fi}. We consider a customer application to a credit described by vector $\hat{x}$, generated from $x$, the true unknown customer application: only the feature "Age in years" has been manipulated, from 47 to 46 in order to allow the applicant to get accepted for the credit by the automated classifier. In order to detect the fraud, the LAD algorithm has been used to generate a list of features that are locally important for the black-box model $b$, which could have been used to manipulate the outcome of $b$. As expected, the age feature has a high feature importance and is the first one a client could easily manipulate. LAD would tell a human expert to check in priority this feature for this illustrative example.

\begin{table}[t]
\centering
\caption{LAD's feature importance for an adversarial example generated on German Credit given a classifier $b$. LAD is able to identify the importance of \textit{Age in years} for the classification}
\begin{tabular}{|c||c|c|c|c|c|c|}
\hline
Features & Credit amount & Credit history & \textbf{Age in years} & Housing & Foreign worker & ... \\ \hline
\begin{tabular}[c]{@{}c@{}}Importance\end{tabular} & 0.43 & 0.28 & \textbf{0.10} & 0.09 & 0.07 & ... \\ \hline
\end{tabular}
\label{table:adverse_fi}
\end{table}
\vspace{-1.5em}

\section{Conclusion and Perspectives}

This paper proposes to discuss a use-case for adversarial defense. It consists in the detection of local adversarial examples for feature-based automated decision problems. Credit insurance provides an example where some people may be tempted to manipulate information they have to provide to increase their chances to get accepted.
While this paper describes a preliminary work, we proposed a method to detect potential adversarial examples by identifying features that have the most impact locally, over the prediction made by a black-box classifier. This method trains an interpretable surrogate to approximate the local decision boundary of the black-box classifier and extract feature importances. These features are provided to a human expert in charge of checking the corresponding information to ascertain the prediction.
A first application on German Credit is used to illustrate our approach: the local fidelity of the method to the classifier it approximates is assessed and the important features to check are extracted for a generated adversarial example.

Our on-going work aims first at consolidating the proposed use-case and formalization of the problem. We plan to improve and test in-depth our proposition (LAD) with a robust experimental protocol, in particular with adversarial examples generated by evasion attack methods to quantify to what extent the approach can help to detect adversarial examples in such context.

\nomenclature{$\hat{x}$}{Potential adverse}
\nomenclature{$b$}{Black-box classifier}
\nomenclature{$\mathcal{X}$}{Feature space domain}
\nomenclature{$X$}{Set of instances $X\in\mathcal{X}$}
\nomenclature{$Y$}{Set of labels $Y = b(X)$}
\nomenclature{$s_{\hat{x}}$}{Local surrogate for $\hat{x}$}
\nomenclature{$X_{sp}$}{Set of support points $x_{sp}^i \in X_{sp}$}
\nomenclature{$X_{seg}^i$}{Set of segment points $x_{seg}^{i,j} \in X_{seg}^i$ drawn on $[\hat{x};x_{sp}^i]$}
\nomenclature{$X_{bt}$}{Set of boundary touchpoints $x_{bt}^k \in X_{bt}$}
\nomenclature{$X_{s_{\hat{x}}}$}{Training set for $s_{\hat{x}}$ drawn in hyperspheres around each $x_{bt}^k$}
\nomenclature{$Y_{s_{\hat{x}}}$}{Set of labels $Y_{s_{\hat{x}}} = b(X_{s_{\hat{x}}})$}
\nomenclature{$\mathcal{F}_{s_{\hat{x}}}$}{Set of most important features from $s_{\hat{x}}$}


%
%
\bibliography{ECML_adv}
\bibliographystyle{bak/splncs03}
\end{document}